# EraseNet : A Recurrent Residual Network for Supervised Document Cleaning

Yashowardhan Shinde [1,2], Kishore Kulkarni [2], Sachin Kuberkar[3]

[1] Department of Computer Engineering, International Institute of Information Technology (I$^2$IT), Pune, India
[2] NxTechWorks, Pune, India
[3] Symbiosis International (Deemed University), Pune, India

*corresponding author (yashowardhanshinde@gmail.com)

**Abstract**

Document denoising is considered one of the most challenging tasks in computer vision. There exist millions of documents that are still to be digitized, but problems like document degradation due to natural and man-made factors make this task very difficult. This paper introduces a supervised approach for cleaning dirty documents using a new fully convolutional auto-encoder architecture. This paper focuses on restoring documents with discrepancies like deformities caused due to aging of a document, creases left on the pages that were xeroxed, random black patches, lightly visible text, etc., and also improving the quality of the image for better optical character recognition system (OCR) performance. Removing noise from scanned documents is a very important step before the documents as this noise can severely affect the performance of an OCR system. The experiments in this paper have shown promising results as the model is able to learn a variety of ordinary as well as unusual noises and rectify them efficiently.

## 1. Introduction

Once printed and out into the world, documents are subject to harsh conditions. Even though the majority of human history is preserved on paper, some of it dates back thousands of years, the information they contain frequently deteriorates extremely quickly. Even in the present era, there are several distortions created that obfuscate the original form of those papers when they are scanned, faxed, copied, and printed. They cannot be recorded in a more permanent, digital form because of their deformations. During the normal course of business, millions of electronic documents, such as invoices and contracts, are evaluated. Scanned documents with various sorts of noise, such as blurred or faded text, salt and pepper (S&P) noise, watermarks, and so on, account for a major portion of them. The performance of OCR and subsequent digitization and analysis is severely hampered by the noise in documents [1, 2]. Enhancing the quality of the documents using image processing methods like denoising and restoration is the first step in automating document analysis. The majority of the literature focuses on eliminating noise from images (e.g., natural landscapes), rather than written documents [1].

However, because of the highly different nature of text documents, these strategies may not be directly relevant. A noise and degradation function may both impair image quality in the image restoration problem. Deblurring, de-fading, and inpainting are some examples. If there is no degradation function in a special situation, the problem is a pure picture denoising problem (e.g., S&P noise removal) [3, 4]. In image processing and computer vision applications spanning from low-level denoising to high-level recognition, deblur, super-resolution, image inpainting, and recovering raw images from compressed images, deep neural networks (DNNs) have proven their superior performance. Despite the progress achieved by DNNs, some problems still exist [3, 4].

Looking at the trends in DNNs, there is a large scope for convolutional neural networks (CNNs) to churn out even better performance on image restoration tasks as CNNs have proven to work extremely well for computer vision problems. Neural networks like Autoencoders [5, 6, 7] and generative adversarial networks (GANs) [8, 9] are generally used for image to image translation tasks like image denoising, image recolouring, image super-resolution, etc. [10, 11]. The proposed network is a Fully Convolutional Network (FCN) [12] that comprises several convolutional and deconvolutional layers along with local residuals in convolutional blocks and symmetrical links between the encoder and decoder, which help in faster training and obtaining higher quality output.

## 2. Related Work

Image denoising being a popular task in computer vision many researchers have developed supervised, self-supervised as well as unsupervised neural networks for denoising, deblurring, and super-resolution tasks. Researchers have employed DNNs, Convolutional networks like Autoencoders [5, 6, 7], GANs [8, 9] and Transformer based networks for solving these tasks [13, 10, 11, 25, 20]. A very deep completely convolutional encoding-decoding framework for image restoration techniques like denoising and super-resolution was proposed by Xiao-Jiao Mao et al. Their network learns end-to-end mappings from corrupted images to the original ones by combining convolution and deconvolution operations [13]. By using networks with a "*blind spot*" in the receptive field, Samuli Laine et al. improved two crucial features, including image quality and training efficiency, without the use of reference data [14]. Instance Normalization was studied by Liangyu Chen et al. in relation to low-level vision tasks. To improve the performance of image restoration networks, they specifically presented the Half Instance Normalization Block (HIN Block)[15]. A straightforward baseline that outperforms SOTA approaches and is computationally effective was developed by Liangyu Chen et al. They demonstrated that the nonlinear activation functions, such as Sigmoid, ReLU, GELU, Softmax, etc., are not required, further simplifying the baseline [16]. The authors of [17, 18] propose a dense U-Net architecture for image denoising tasks with residual connects between the encoder and decoder similar to the RedNet architecture proposed by Jiang et al. for semantic segmentation [19]. A self-supervised learning technique was proposed by Yuhui Quan et al. that employs only the input noisy image itself for training. The network is trained with dropout using pairs of Bernoulli-sampled examples of the input image in the proposed technique [20]. By modifying many crucial building pieces (multi-head attention and feed-forward network) such that it may capture far-reaching pixel interactions while still being suitable to huge images, Syed Waqas Zamir et al. suggested an effective Transformer model [21]. With an emphasis on issues related to image denoising, Stanley Osher et al. presented a new iterative regularisation method for inverse problems based on the usage of Bregman distances [22]. Zhenwen Dai et al. attempted to automatically remove corruptions from a single letter-size page. Their method begins by autonomously learning character representations from document patches. They parameterized pattern features, their planar layouts, and their variances in a probabilistic generative model for learning [23]. In order to help the reader have a thorough knowledge of how various strategies relate to one another, Peyman Milanfar provided a useful and approachable framework for comprehending some of the fundamental principles behind some of the image denoising techniques. Additionally, he draws comparisons between these methods and more traditional (empirical) Bayesian strategies [24]. Image Super Resolution architectures like the SRResNet, SRGAN [10] that used a PixelShuffle layer for upsampling, EDSR, MDSR, and WDSR [11] can be used for image denoising tasks. With respect to the upscaling procedure, Wanjie Sun et al. suggested a learning image downscaling approach based on content adaptive resampler (CAR) [25]. They suggested creating a resampler network that would produce content-adaptive picture resampling kernels that would be applied to the original, high-resolution input to produce pixels on the downscaled image. An unsupervised document cleaning approach has been explored by Gangeh et al. [1] that can deal with noise like, watermarks, salt & pepper noise, as well as blurred and/or faded text from documents. This approach integrates a deep mixture of experts with cycle-GAN [26] as the base network.

## 3. Implementation Details

## 3.1 Model Architecture

The proposed architecture is fully convolutional and deconvolutional. The architecture can be split into 2 parts: the encoder and the decoder, with skip connections from encoder to decoder. The aim of the encoder is to compress all the important information or features of the document into a latent vector that can be then used to reconstruct the original image with the help of the deconvolutional decoder. The intuition behind this kind of architecture is that during the encoding phase, the encoder can compress the image such that only the important features, excluding the noise, will be captured into the latent vector. Thus, during the decoding phase the image will be reconstructed, eliminating different noises in the document.

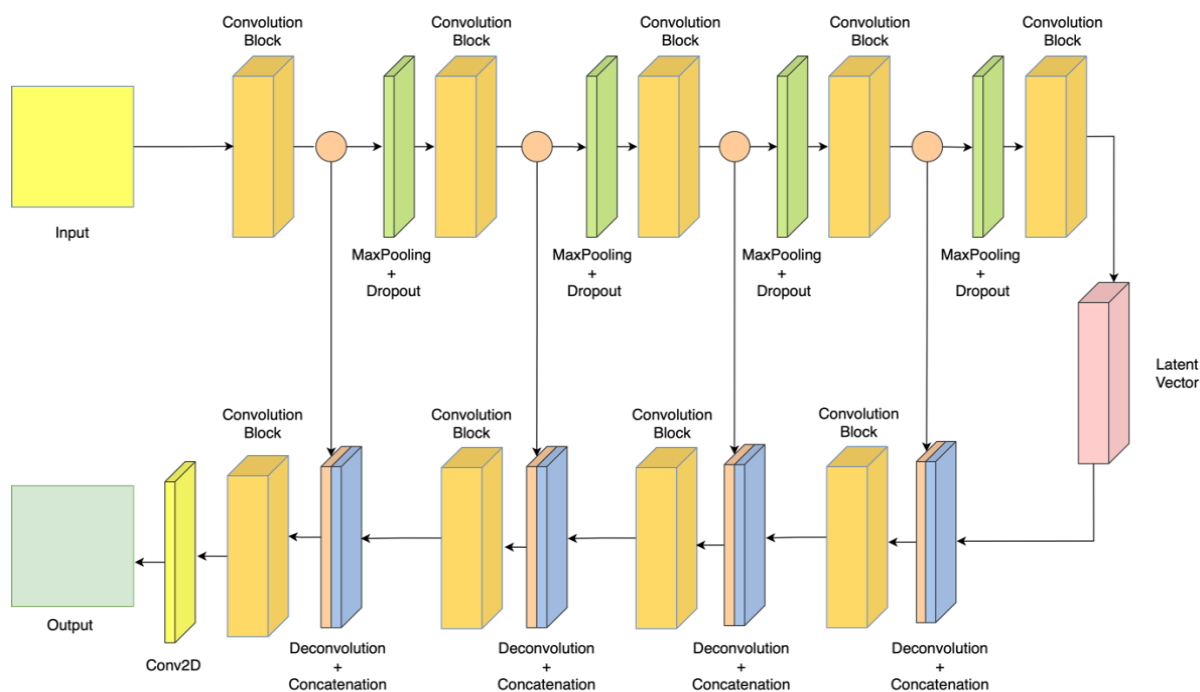

Figure 1. Architecture of the proposed EraseNet network. The part above the latent vector represents the encoder and the part to below represents the decoder with symmetric skip connections present between the encoder and the decoder.

### 3.1.1 Convolution Block:

The convolutional block is built using a combination of residual fission architectures proposed by DenseNet [27] and ResNet [28]. The DenseNet [27] proposed a residual network where the residual is concatenated to the output of its subsequent layers, whereas ResNet [28] proposed an architecture where the residual is added to the output of its subsequent layer. The structure of the convolution block can be observed in Figure 2.

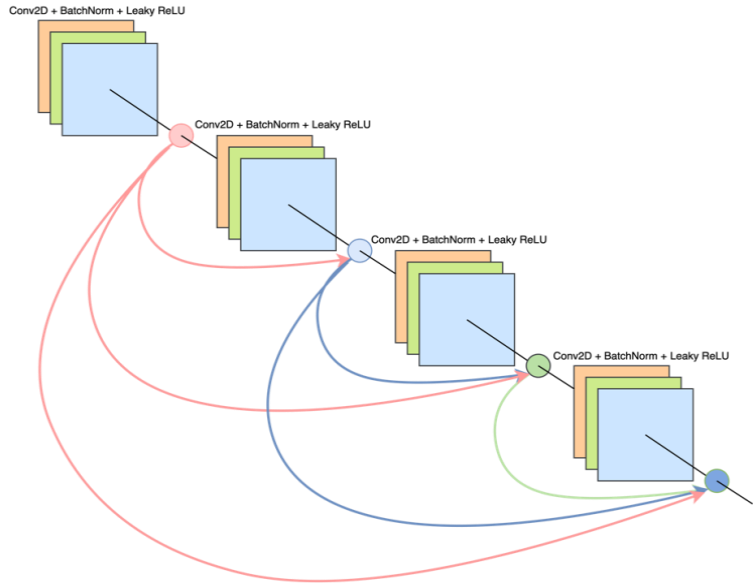

Figure 2. Proposed convolution block with recurrent residual connections which is a building block of the EraseNet Network.

In this paper, addition of residuals is preferred over concatenation as a method of fission because it makes larger feature maps feasible, as opposed to the problem of the feature-map growth discussed in DenseNet [27]. This type of architecture ensures that there is minimum information loss by introducing direct connections from any layer to all subsequent layers. As a result, the $l$th layer receives the features of all preceding layers, $x_0, \ldots, x_{(l-1)}$, as input. Formula:

$$x_l = F([x_0, x_1, x_2, \ldots, x_{l-1}]) \quad (1)$$

where $[x_0, x_1, x_2, \ldots, x_{l-1}]$ refers to the addition of the feature-maps produced in layers $0, 1, \ldots, l-1$.

### 3.1.2 Composite Function:

Motivated by ResNet [28] the composite function in the ConvBlock is defined by a series of 3 consecutive operations: a $(3 \times 3)$ or $(5 \times 5)$ convolution (Conv) followed by batch normalization (BN), and leaky rectified linear unit (Leaky ReLU) (alpha=0.2) [29]. The Leaky ReLU activation is defined as :

$$y_i = \begin{cases} x_i & if\ x_i \geq 0 \\ \frac{x_i}{\alpha} & if\ x_i < 0 \end{cases} \quad (2)$$

where $\propto$ is a fixed parameter in range (1, +∞). There are a few advantages to using Leaky ReLU over ReLU: Due to the absence of zero-slope portions, it resolves the **dying ReLU** [30] issue. It speeds up training. According to [29], training is faster when the mean activation is close to 0. Leaky ReLU is more balanced than ReLU, which may lead to faster learning.

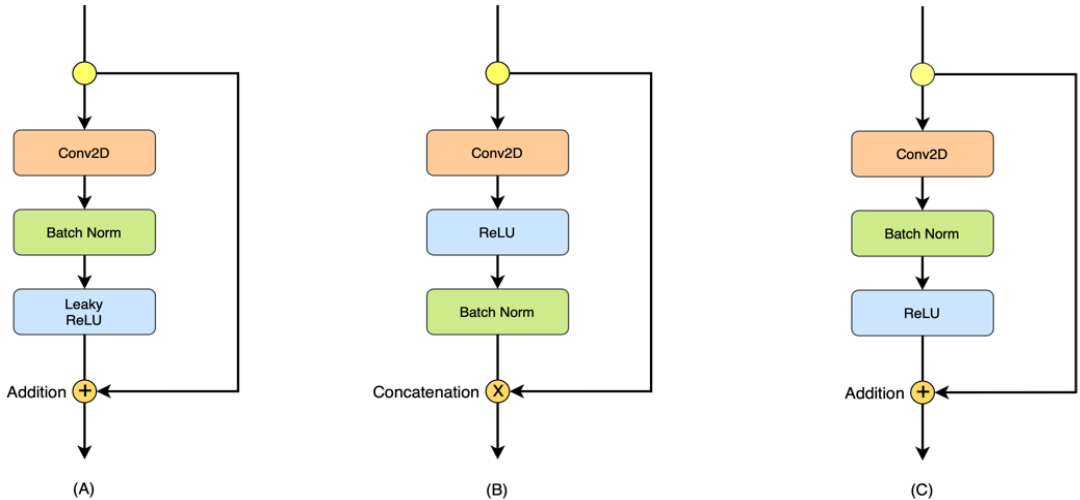

Figure 3. Comparison between the proposed composite function, DenseNet and ResNet. (A) represents the proposed composite function, (B) represents the composite function of DenseNet and (C) represents the composite function of ResNet.

### 3.1.3 Encoder

The aim of the encoder is to compress all the important information in the document without capturing the noise into a latent vector, which can be used to reconstruct the document, eliminating the noise in the process. The encoder is constructed by stacking a series of convolutional blocks followed by the transition layers for compressing the image. The transition layer is defined as a series of 2 operations: a MaxPooling followed by Dropout layer. For the MaxPooling layers, the kernel size is set to (2x2) and the for the Dropout layers, the dropout factor is set to (0.3). The residuals are stored before each transition layer. These residuals are then used for skip connections between the encoder and the decoder.

### 3.1.4 Decoder

The main aim of the decoder is to reconstruct the original image from the latent vector generated by the encoder while at the same time eliminating the noise during the reconstruction process. The decoder is constructed using deconvolution [31, 32] layers with rectified linear unit (ReLU) activation followed by feature fission and a convolutional block. Instead of an upsampling layer, which is a simple layer with no weights that doubles the dimensions of input, the decoder employs a Transpose Convolutional (Deconvolution) [31, 32] layer, which is an inverse convolutional layer that upsamples the input and learns how to fill in details throughout the model training process [31, 32, 33]. The symmetric skip connections between the encoder and decoder use concatenation as the method for fission. As the shape of the image has to be the same for concatenation, the fission is performed after upsampling the image using deconvolution layers. Consequently, fission can be represented using the formula:

$$x_i = H([r_j, x_i]) \qquad (3)$$

Here, $x_i$ is the output of the $i^{th}$ deconvolutional layer and $r_j$ is the $j^{th}$ residual which has the same size as $x_i$. The number of output channels might differ here. Finally, the last layer of the network is a Conv2D layer with 1 channel and a sigmoid activation function. The architecture of EraseNet can be seen in Table 1.

Table 1. Architecture of the proposed EraseNet Network with 4 down-sampling and 4 upsampling operations

| Layers | Output Size | EraseNet |
|---|---|---|
| ConvBlock (1) | 256x256 | ConvBlock(Depth = 2, Filters = 64, Kernel Size = (5x5)) |
| Residual (1) | 256x256 | |
| Transition Layer (1) | 128x128 | MaxPooling2D (2x2) |
| | | Dropout (0.3) |
| ConvBlock (2) | 128x128 | ConvBlock(Depth = 2, Filters = 64, Kernel Size = (5x5)) |
| Residual (2) | 128x128 | |
| Transition Layer (2) | 64x64 | MaxPooling2D (2x2) |
| | | Dropout (0.3) |
| ConvBlock (3) | 64x64 | ConvBlock(Depth = 3, Filters = 128, Kernel Size = (3x3)) |
| Residual (3) | 64x64 | |
| Transition Layer (3) | 32x32 | MaxPooling2D (2x2) |
| | | Dropout (0.3) |
| ConvBlock (4) | 32x32 | ConvBlock(Depth = 3, Filters = 256, Kernel Size = (3x3)) |
| Residual (4) | 32x32 | |
| Transition Layer (4) | 16x16 | MaxPooling2D (2x2) |
| | | Dropout (0.3) |
| ConvBlock (5) | 16x16 | ConvBlock(Depth = 4, Filters = 512, Kernel Size = (3x3)) |
| ConvTranspose (1) | 32x32 | Conv2DTranspose(Filters = 256, Kernel Size = (3x3), Stride = 2) |
| Concatenation (1) | 32x32 | Concatenate( ConvTranspose (1), Residual (4)) |
| ConvBlock (6) | 32x32 | ConvBlock(Depth = 3, Filters = 256, Kernel Size = (3x3)) |
| ConvTranspose (2) | 64x64 | Conv2DTranspose(Filters = 128, Kernel Size = (3x3), Stride = 2) |
| Concatenation (2) | 64x64 | Concatenate(ConvTranspose (2), Residual (3)) |
| ConvBlock (7) | 64x64 | ConvBlock(Depth = 3, Filters = 128, Kernel Size = (3x3)) |
| ConvTranspose (3) | 128x128 | Conv2DTranspose(Filters = 64, Kernel Size = (3x3), Stride = 2) |
| Concatenation (3) | 128x128 | Concatenate(ConvTranspose (3), Residual (2)) |
| ConvBlock (8) | 128x128 | ConvBlock(Depth = 3, Filters = 64, Kernel Size = (5x5)) |
| ConvTranspose (4) | 256x256 | Conv2DTranspose(Filters = 64, Kernel Size = (3x3), Stride = 2) |
| Concatenation (4) | 256x256 | Concatenate(ConvTranspose (4), Residual (1)) |
| ConvBlock (9) | 256x256 | ConvBlock(Depth = 3, Filters = 64, Kernel Size = (5x5)) |
| Convolution (1) | 256x256 | Conv2D(Filters = 1, Kernel Size = (3x3)) |
| Sigmoid (1) | 256x256 | Sigmoid |

## 3.2 Training

In order to train the network for mapping the noisy images to the clean images, it is important to estimate the convolutional and deconvolutional kernel weights (Θ). This is accomplished by reducing the Euclidean loss between the network's outputs and the clean image. Given N training sample pairs $(X_i, Y_i)$ where, $X_i$ is the corrupted version of the image and $Y_i$ is the ground truth, the Mean Squared Error (MSE) is defined as:

$$\mathcal{L}(\Theta) = \frac{1}{n} \sum_{n=1}^{N} \|F(X_i, \Theta) - Y_i\|_F^2 \qquad (4)$$

Here, $F(X_i, \Theta)$ represents output of the network. The Adam optimizer [13] was used for training the network with an initial learning rate of $10^{-4}$ as Adam was found to converge more quickly than conventional stochastic gradient descent (SGD) in practise. Unlike [34, 35], where a lower learning rate is set for the final layer, the base learning rate is kept uniform for all layers as this approach is not necessary in the proposed network. Models frequently gain by reducing the learning rate after learning reaches a plateau, hence the *ReduceLROnPlateau* learning rate scheduler is used for training [36, 37]. This callback tracks *validation loss* and reduces the learning rate by 0.1 if no progress is observed after a *patience* number of epochs that is set to 10. The implementation and training of the network is done using TensorFlow.

Grayscale images of size (256x256) and (864x480) are used for training the network. The patches of size (256x256) have been extracted from the documents by converting each document page into grayscale, resizing them to (1024x768) and extracting 12 patches of (256x256) from the page. A sliding window technique is used to accomplish this, moving from the top left corner to the bottom right corner. In practice, it was observed that using the proposed method for patch extraction works better than extracting the patches randomly. The (864x480) images are basically the original pages resized into the required shape.

## 4. Experiments and Results

### 4.1 Datasets

The proposed model EraseNet is evaluated on two datasets. **Dataset 1,** also known as the NoisyOffice dataset, was created by RM.J. Castro Bleda, et al. and is hosted by the UCI machine learning repository and is also available on Kaggle [38]. This dataset consists of 144 training and testing images of pages with noise and their corresponding cleaned outputs.

**Dataset 2** is a custom dataset created in house using the scanned pages from a collection of available legal documents that had to be restored and digitized. This dataset is a collection of 660 images along with their corresponding cleaned output. These documents contain a wide variety of noises, like black residuals left after scanning documents, torn page marks, oil stains, creases, etc. Due to this wide range of noises, around 50 to 60 images are collected for each kind of noise to make this dataset.

### 4.2 Evaluation and Results

For each of the datasets, Structural Similarity Index (SSIM) [39, 40] and Peak Signal-to-Noise Ratio (PSNR) [39, 40] are calculated for evaluation. PSNR and SSIM can be represented using equations 5 and 6 respectively.

$$PSNR = 20 * log\left(\frac{MAX_I}{\sqrt{MSE}}\right) \quad (5)$$

Here, MAX represents the maximum value in the distribution and MSE is the Mean Squared Error defined in equation 3.

$$SSIM(x, y) = \frac{(2\mu_x\mu_y + c_1)(2\sigma_{xy} + c_2)}{(\mu_x^2 + \mu_y^2 + c_1)(\sigma_x^2 + \sigma_y^2 + c_2)} \quad (6)$$

Here, x and y represent the two images, $\sigma_x^2$ and $\sigma_y^2$ represent the variance of x and y respectively, $\mu_x$ $and$ $\mu_y$ are the mean of x and y respectively, $\sigma_{xy}$ is the covariance of x and y. $C_1$ and $C_2$ are parameters used to stabilize the division with a small denominator.

For the proposed method denoted as EraseNet, two versions are implemented: EraseNet-3, containing 3 down-sampling and 3 upsampling operations; and EraseNet-4 containing 4 down-sampling and 4 upsampling operations. The models were trained for 100 epochs on the Dataset 1 and for 250 epochs on Dataset 2. The results of these 2 networks have been shown in Table 2.

Table 2. Performance comparison between architectures of Erasenet-3 and EraseNet-4 on NoisyOffice dataset and the custom dataset (PSNR(dB), SSIM, MSE). All the scores mentioned are average scores calculated over the test images in the datasets.

| Dataset | Model | PSNR(dB) | SSIM | MSE |
|---|---|---|---|---|
| NoisyOffice | EraseNet-3 | **83.33** | **0.9878** | **3.02e-4** |
| | EraseNet-4 | 83.14 | 0.9766 | 3.155e-4 |
| Custom Dataset | EraseNet-3 | 35.38 | 0.8524 | 18.83 |
| | EraseNet-4 | **36.36** | **0.8892** | **15.39** |

As observed in Table 2, the EraseNet-4 gives a better performance on the Datasets 2 with a PSNR of 36.36, and a SSIM score of 0.8892, and the EraseNet-3 performs better on the Dataset 1 with a PSNR of 83.33 and a SSIM score of 0.9878.

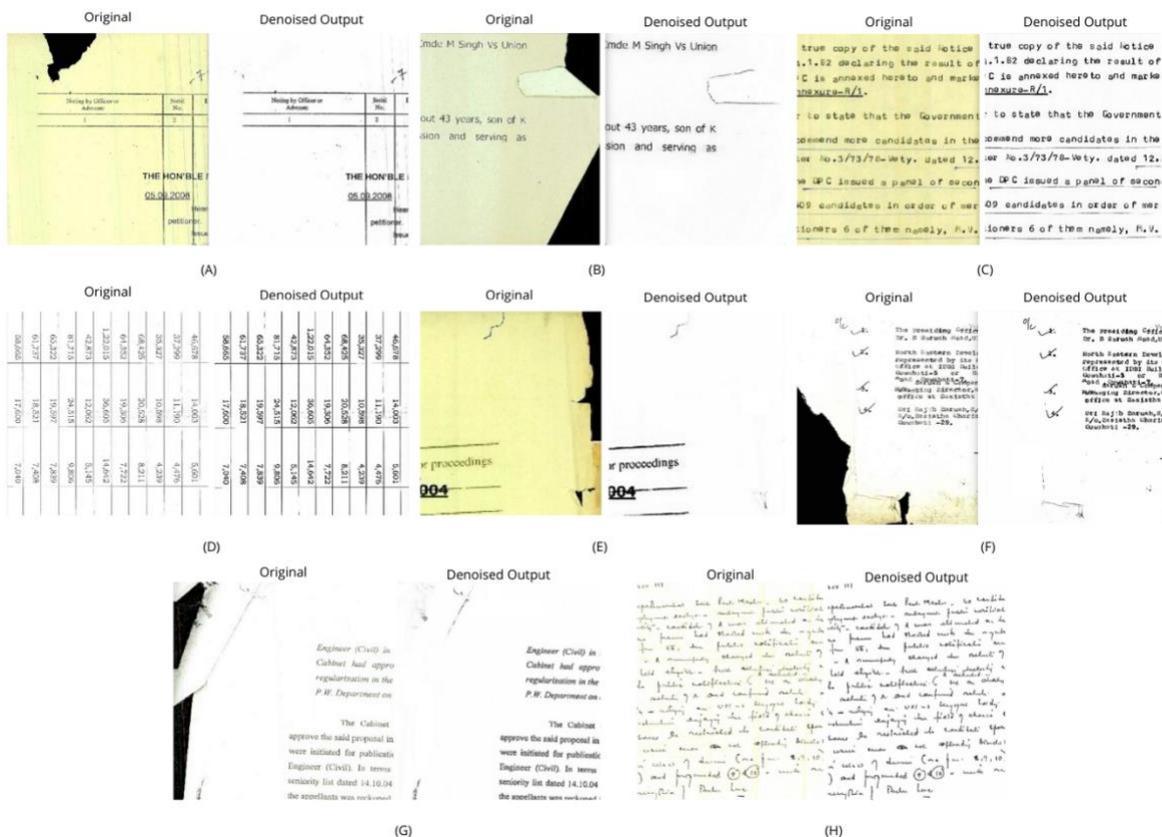

Figure 4. Results of the EraseNet-4 on randomly selected test images from the Dataset 2. Samples (A), (B), (E), (F) show noises like torn or folded pages, (C), (D), (G) and (H) show noises like faded hand-writing and text.

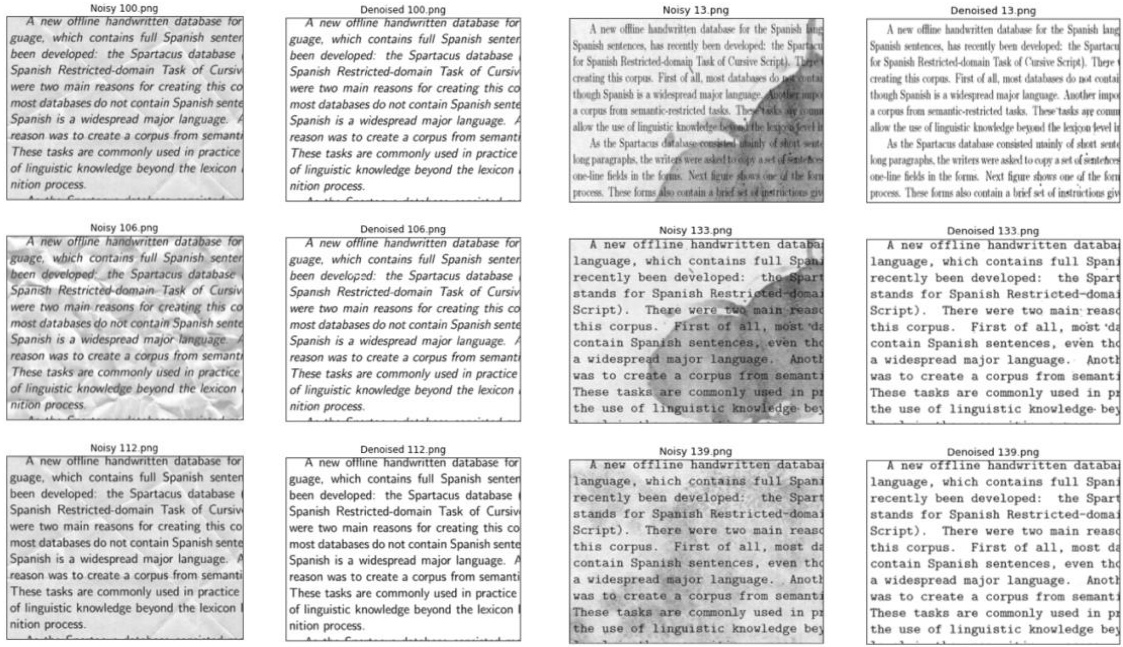

Figure 5. Results of the EraseNet-3 on randomly selected test images from the Dataset 1.

It can be observed from Figure 4 and Figure 5, that the models are able to remove most of the noise from the pages while retaining the important details of the page. The deformities such as torn pages, ink or oil stains, crease marks, and light text are completely eliminated using the proposed model.

The network can clean document images of any size. Given a test image, one can simply pass the image through the network, which is capable of producing great results. In order to achieve a more sharp and crisp output, the result can be further processed using a 2D sharpness filter, which helps in obtaining a more sharp and crisp result, as observed in practice. The sharpening kernel is defined as:

$$kernel = \begin{bmatrix} [0, & -1, & 0], \\ [-1, & 5, & -1], \\ [0, & -1, & 0] \end{bmatrix}$$

Another practice observed in [4] for achieving better results is passing the image through the network in different orientations and then taking the average of the results. However, in practice, this method increases the inferencing time significantly and fails to demonstrate any substantial enhancement in the image quality.

## 5. Conclusion and Future Work

A deep encoding and decoding system for document restoration has been proposed in this research. Combining convolution and deconvolution, the restoration problem is modelled by removing the main image information and recovering details. In addition, the presented network makes use of both global and local skip connections, which contributes in the restoration of clear images and address the optimization challenge provided by the gradient vanishing problem, resulting in the best performance. The study and the results of the experiments performed demonstrate that the suggested network performs really well when it comes to document denoising. The same model architecture can be used for solving problems like image denoising tasks like (S&P) noise, deblurring, de-raining, etc. As the data for supervised models is not easily available, an unsupervised approach for cleaning documents can be a topic of further research. Researchers can also explore training the proposed model in an unsupervised way similar to GANs.